%% file: lightfieldEdit.tex
\documentclass[journal,comsoc]{IEEEtran}
\usepackage[T1]{fontenc}
\usepackage{amsmath}
\usepackage[cmintegrals]{newtxmath}
\usepackage{times}
\usepackage{epsfig}
\usepackage{graphicx}
\usepackage{enumitem}
\usepackage{subfigure}
\usepackage{setspace}
\usepackage{multirow}
\usepackage{multicol}
\usepackage{url}
\usepackage{caption}
\usepackage{wrapfig}
\usepackage{color}

\def \u {\mathbf{u}}
\def \x {\mathbf{x}}
\def \view {\mathbf{I}}
\def \lf {\mathbf{L}}
\def \loss {\mathcal{L}}

\begin{document}
\title{A Learned Compact and Editable Light Field Representation}

\author{Menghan~Xia, Jose~Echevarria, Minshan~Xie, and~Tien-Tsin~Wong
\thanks{Menghan Xia, Minshan Xie and Tien-Tsin Wong are with the Department of Computer Science and Engineering, The Chinese University of Hong Kong, HK. \{mhxia, msxie, ttwong\}@cse.cuhk.edu.hk.}
\thanks{ Jose Echevarria is with the Adobe System Inc, US. echevarr@adobe.com.}
}


\maketitle

\begin{abstract}
Light fields are 4D scene representation typically structured as arrays of views, or several directional samples per pixel in a single view. This highly correlated structure is not very efficient to transmit and manipulate (especially for editing), though.
To tackle these problems, we present a novel compact and editable light field representation, consisting of a set of visual channels (i.e. the central RGB view) and a complementary meta channel that encodes the residual geometric and appearance information. The visual channels in this representation can be edited using existing 2D image editing tools, before accurately reconstructing the whole edited light field back. 
We propose to learn this representation via an autoencoder framework, consisting of an encoder for learning the representation, and a decoder for reconstructing the light field. To handle the challenging occlusions and propagation of edits, we specifically designed an editing-aware decoding network and its associated training strategy, so that the edits to the visual channels can be consistently propagated to the whole light field upon reconstruction.
Experimental results show that our proposed method outperforms related existing methods in reconstruction accuracy, and achieves visually pleasant performance in editing propagation.
\end{abstract}


\IEEEpeerreviewmaketitle

\input{Introduction}
\input{RelatedWorks}
\input{Method}
\input{Results}

\section{Conclusion}
\label{sec:conclusion}

We presented a new compact and editable light field representation, which enables light fields to be transferred and edited robustly in existing 2D tools and pipelines effortlessly.
This efficient representation is learned through a special decoding process, split into specific subnetworks that ease the training of our model. 
Experimental results show the effectiveness of the proposed method, both in terms of reconstruction accuracy and quality of the propagated edits, comparable to previous work.
Obvious next steps are geometry-aware edits that may require a combination of edits and updates to both visual and meta channels, or extensions to light field videos.
We believe this representation opens news directions for light field processing and editing, and we hope it inspires follow-up works in this area. 

%
%
\bibliographystyle{IEEEtran}
\bibliography{lightfieldEdit}

\end{document}

%% file: Introduction.tex
\section{Introduction}
\label{sec:introduction}


\IEEEPARstart{4}{D} Light fields model incoming light rays hitting the camera sensor at different locations and from different angles, allowing them to capture complex geometries and material appearances faithfully~\cite{LevoySIG96,GortlerSIG96}. This rich representation has applications in synthesis of novel viewpoints and refocusing~\cite{NgCSTP05}, or geometry~\cite{WangECCV16} and material analysis~\cite{AlperovichCVPR18}.
Given a raw light field is typically structured as an array of multiple images from different viewpoints, or multiple directional samples per pixel, such memory-intensive collections of rays are challenging to handle or edit efficiently. 
Several compression~\cite{ChenTIP18,MiandjiSIG19} and editing techniques~\cite{JaraboTOG14,ChenICME15,ZhangTVCG17} have been proposed separately.
However, although the latter enable new kinds of edits only possible with light fields, the amount of highly correlated data needed in run-time makes naive approaches inefficient, and the proposed tools are too disruptive for typical edits, with respect to the established 2D image editing workflows. So, a more memory-efficient data structure that is easier to digest by existing tools would be desirable.

\begin{figure}[!t] 
	\centering
	\includegraphics[width=1\linewidth]{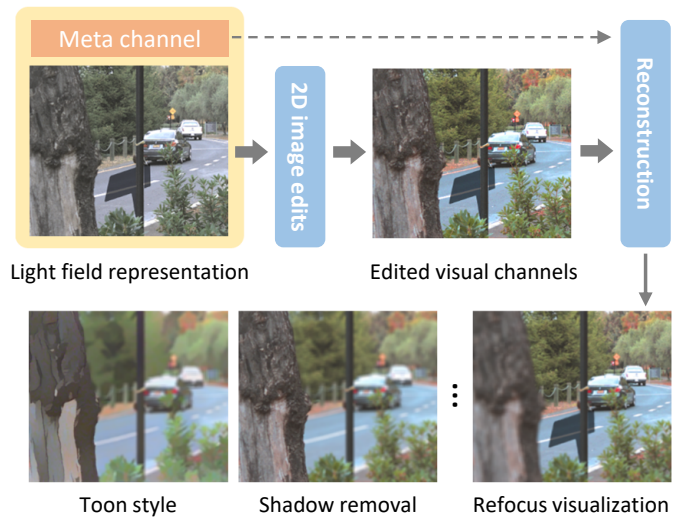}
	\caption{Our proposed representation enables the light filed edits to be no more than a 2D image edits, which is memory-efficient and supports most image processing algorithms (e.g. color transfer, style transfer, shadow removal, etc) naively. Moreover, due to its compactness, the process can be performed in lightweight mobile devices or even via online cloud computation.}
 	\label{fig:teaser}
\end{figure} 

To tackle this problem, we propose to represent the light field as a single-view multi-channel image, which consists of visual channels that capture the visual content of the scene (i.e. the central RGB view) and a complementary meta channel that stores other view-dependent information like geometric or material properties. This meta channel would allow to reconstruct the original light field deterministically. Obviously, this representation is compact thanks to the efficient information encoding in the meta channel. Furthermore, we require the meta channel to be compatible to edits to the visual channels, such that the editing effects on the visual channels can be consistently propagated to the other views when the full light field is reconstructed. In practice, this two features make our representation suitable for lightweight desktop or mobile editing interface, which requires memory no more than a typical 2D RGBA image and supports popular 2D image processing algorithms natively (Fig.~\ref{fig:teaser}).
We believe it makes an importance step to promote the light field consumption at scale, as currently prevented by large bandwidth and specialized complex editing tools.

To construct such a representation, we propose to learn the encoding-decoding scheme in data-driven manner.
Inspired by Xia et al.~\cite{XiaSIGA18}, we adopt an autoencoder framework (Fig.~\ref{fig:overview}). Specifically, the encoder subnetwork converts a light field into a single meta channel that, together with the reference RGB view, forms the representation that allows the decoder subnetwork to recover the full light field any time.
The two subnetworks are then jointly trained to facilitate the invertibility of the learned representation of the original light field.
However,  it is non-trivial to reconstruct a desired light field from a possibly edited representation. 
To facilitate the edit-aware reconstruction, we specifically decompose the decoding process into three modules: feature separation that extracts individual view information from the meta channel; disparity recovery that warps the visual channels to other views; fusion synthesis that further restores the view-dependent visual content from the corresponding feature map and then synthesize the target view based on the warped result. Note that the fusion synthesis module not only integrates the information encoded from the input, but predicts the potential editing effects of the occlusion regions based on the surrounding context.
We train our model using publicly available light field datasets to learn the invertible representation under self-supervision. Additionally, we introduce several edits to the visual channels during training, in order to enforce the editing propagation functionality.

To verify the feasibility of our approach, we applied our method on several light fields from publicly available datasets, which were captured with Lytro cameras under various settings. Extensive quantitative and qualitative evaluations were conducted. Results show that our method achieved high reconstruction accuracy and consistent editing propagation effects. In summary, our main contributions include:
\setitemize[1]{itemsep=1pt,partopsep=0pt,parsep=\parskip,topsep=3pt}
\begin{itemize}
	\item We present a novel light field representation that is both compact and editable. It allows low-bandwidth transmitting and efficient visual edits using standard 2D image editing tools, with the complex geometric properties preserved.
	\item We propose an effective method to learn such representation and its reconstruction, which achieves quality representation accuracy and edit propagation.
	\item Our method opens a promising direction for handy light field processing, which could promote light field applications to massive use.
\end{itemize}

%% file: RelatedWorks.tex
\section{Related Works}
\label{sec:related_works}

 
\subsection{Light Field Reconstruction}
\label{subsec:lightfield_reconstruction}

Previous works on light field reconstruction focus on synthesizing a dense light field from a sparser set of samples. Traditional methods mainly formulate the view synthesis as an optimization problem to generate the novel views directly~\cite{ShiTOG14,VagharshakyanICIP15} or estimate the disparity to obtain them through warping~\cite{WannerTPAMI14,LiTIP15}. These methods tend to present ghosting and tearing artifacts when the input views are sparse.
Recently, deep learning techniques have been explored for synthesis of dense views~\cite{GuoACCV16,YoonICCVW15}. Kalantari et al.~\cite{KalantariTOG16} synthesize a novel view with two sequential CNNs that estimate disparity for warping and refine the colors jointly. Wu et al.~\cite{WuCVPR17} focus on recovering the high frequency details of linearly upsampled epipolar-plane images (EPIs), where a blur-deblur scheme is employed to tackle the information asymmetry problem. To deeply characterize spatial-angular clues, Yeung et. al.~\cite{WingECCV18} employ spatial-angular alternating convolutions within a residual learning framework. Different from these methods requiring supervision of dense light fields, Ni et al.~\cite{NiCGF19} utilize inter-view cycle consistency to enable the unsupervised learning of a light field from two input views, where occlusions are compensated through a forward-backward warping scheme. Reducing the amount of input views to one, Srinivasan et al.~\cite{SrinivasanICCV17} explores the task of synthesizing a light field from a single image, achieving impressive results for very specific scenes. In all of them, the ill-posed sparse-to-dense reconstruction struggles to generate or recover information invisible in the input views due to occlusions. 

A different direction for light field reconstruction uses single or few coded images~\cite{MarwahTOG13,VadathyaACPR17,InagakiECCV18}, which as a compact representation at data capturing targets for accurate reconstructions. Anyhow, such coded images are usually generated via hand-crafted procedures, which significantly limits their expressiveness and accuracy; and using the visual channels for the encoding limits their editability afterwards.
More recently, multiplane images (MPIs) have been also proposed as an effective representation for light fields~\cite{ZhouTOG18,FlynnCVPR19,SrinivasanCVPR19}, with very efficient generation of new views for scenes with complex geometries and materials. However, it is not clear how to edit such overabundance of layers that, although effective for parallax effects, do not seem to model the implicit geometries and materials faithfully.

\subsection{Light Field Editing}
\label{subsec:lightfield_editing}

Most previous works focus on specific editing tasks such as retargeting~\cite{BirklbauerCGF12}, morphing~\cite{WangTVCG05} or completion~\cite{PenduTIP18}. These specialized editing methods are difficult to generalize and unify in a single framework, so users cannot perform multiple typical edits with the same tool. 
A different line of research aims for more general edits instead. Seitz et al.~\cite{SeitzIJCV02} propagate local edits across multiple views through a voxel-based light field representation. Jarabo et al.~\cite{JaraboSIACG15} propose a novel downsampling-upsampling propagation scheme to propagate sparse edits to the full light field based on an affinity function. 
Chen et al.~\cite{ChenICME15} extend the versatile patch-based editing framework to the domain of light fields and enable several interesting new editing operations.
Zhang et al.~\cite{ZhangTVCG17} decompose the central view into a set of layers at different depths, so existing patch-based edits can be applied to them.
Recently, Beigpour et al.~\cite{Beigpour18} present an intrinsic decomposition framework for editing the appearance of surfaces through various band shift operators.
Aiming to exploit the 4D information in a light field, Jarabo et al.~\cite{JaraboTOG14} propose and study novel interfaces and workflows.
Different from all the methods above, we propose a new compact editable light field representation that enables light fields to be edited efficiently using existing 2D image editing tools.

%% file: Method.tex
\section{Compact Editable Light Fields}
\label{sec:approach}

Given a 4D light field, we aim to represent it as a single meta channel and several visual channels (like an RGB image), which in the one hand is apparently compact, and in the other, the edits on the visual channels can be propagated consistently to the rest of the views whenever the full light field is reconstructed.

%
\begin{figure}[!t]
	\centering
	\includegraphics[width=1\linewidth]{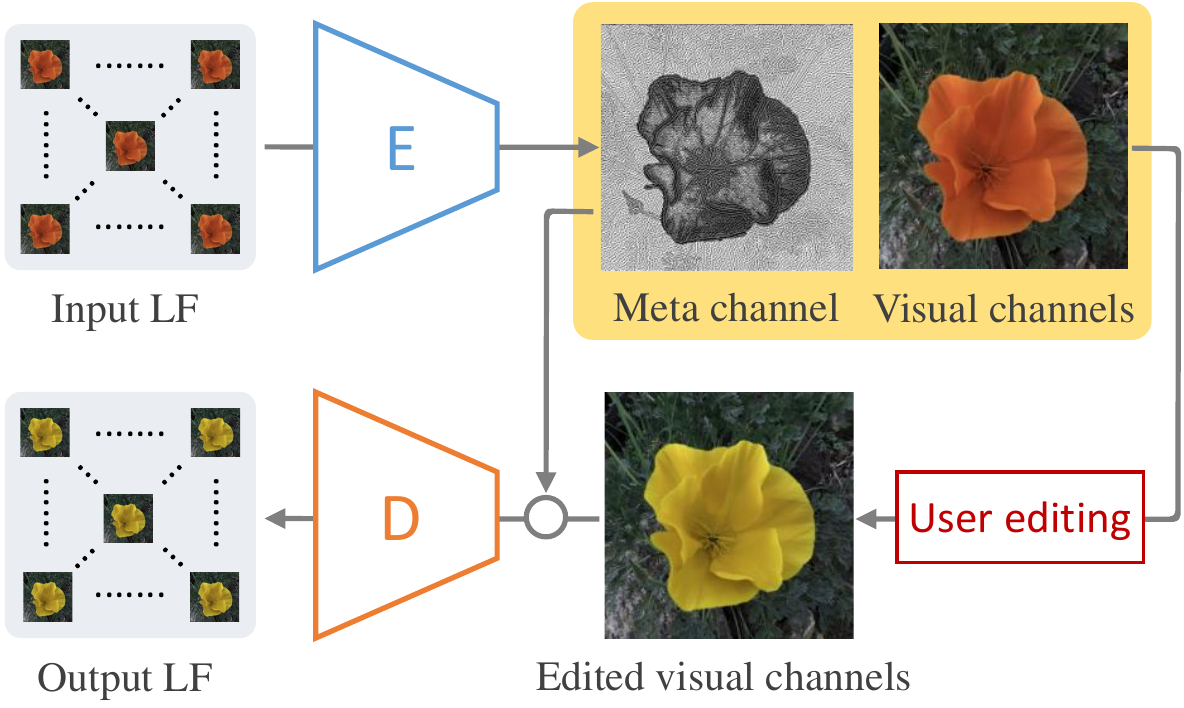} 
	\caption{System overview. Given an input light field, the encoder converts it into a single meta channel, which together with the visual channels (the central RGB view) serves as a compact and editable representation of the original light field. Users can edit the appearance of the visual channels using standard 2D editing tools. Then the edited visual channels and original meta channel are taken by the decoder to reconstruct a full light field with all the edits consistently propagated.}
 	\label{fig:overview}
\end{figure}

\subsection{Problem Formulation}
\label{subsec:framework}

We denote a structured 4D light field consisting of an array of $M\times N$ RGB views as $\lf=\{\view_i\}_{i=1}^{M\times N}$, where $\lf(\u,\x)$ samples the light field at angular coordinate $\u=(u,v)$ and spatial coordinate $\x=(x,y)$, and $\view_i=\lf(\u_i)$. 
As illustrated in Figure~\ref{fig:overview}, our framework consists of an encoding subnetwork (encoder) $\mathcal{E}$ and a decoding subnetwork (decoder) $\mathcal{D}$.
The encoder takes a light field $\lf$ as input, and generates a meta channel $\mathbf{Z}$ which is a float-valued 2D array of the same resolution as $\mathbf{I}_i$. 
We construct a tuple $\{\mathbf{Z}, \view_c\}$ as the intermediate representation, where $\view_c$ denotes the editable central view of $\lf$. 
Inversely, the decoder $\mathcal{D}$ reconstructs a light field $\tilde{\lf}=\{\tilde{\view}_i\}_{i=1}^{M\times N}$ from the edited pair $\{\mathbf{Z}, \tilde{\view}_c\}$, where $\tilde{\view}_c$ is the edited version of $\view_c$.  Formally, they can be described as:
\begin{eqnarray}
\label{eq:formulation}
\mathbf{Z}  &=& \mathcal{E}(\,\,\lf\,)  \\
\tilde{\lf} &=& \mathcal{D}(\,\mathbf{Z}, \tilde{\view}_c\,)
\end{eqnarray}
where $\mathcal{E}$ and $\mathcal{D}$ roughly work as a pair of inverse functions with their own trainable parameters. 
According to our goal, the reconstructed $\tilde{\lf}$ should be as similar as possible to a target light field $\hat{\lf}$, whose appearance is coherent with $\tilde{\view}_c$, while maintaining the same scene geometry of $\lf$. 
This process becomes very challenging because of (dis)occlusions, since the meta channel is encoded from the original light field, independently from the potential edits to the visual channels. As a starting point, $\tilde{\view}_c=\view_c$ implies no editing involved, so we have $\hat{\lf}=\lf$, which enables self-supervised learning. In Section~\ref{subsec:edit_reconstruction}, we will discuss how to train our model with edits propagation considered.

\subsection{Representation Space}
\label{subsec:embeding_space}

%
\begin{figure}[!t]
	\centering
	\includegraphics[width=1.0\linewidth]{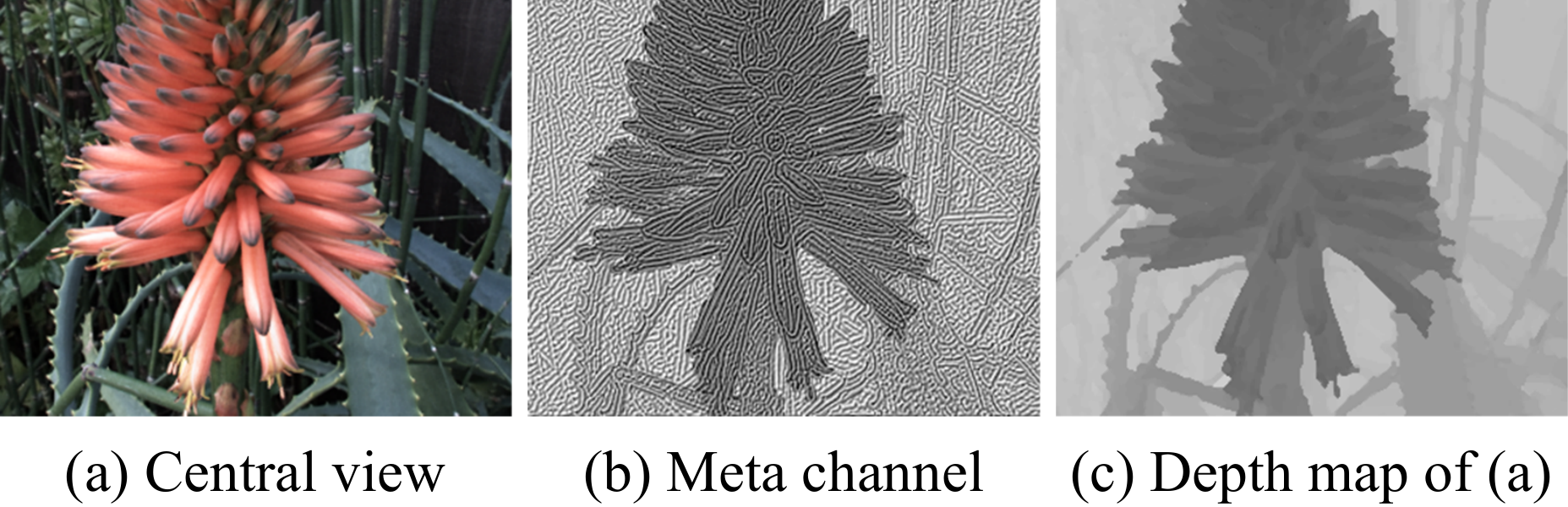}
	\caption{Representation visualization: Visual channels, i.e. the central view (a); Meta channel (b) that roughly resembles the depth distribution of the central view, with some extra structured patterns. The depth map  (c) for this light field was estimated using~\cite{WangICCV15}.}
 	\label{fig:representation}
\end{figure}

Our model learns to represent a light field as a single meta channel paired with the central RGB view. The central view  captures the reference visual content of the light field, so the encoder only needs to learn a complementary residual map, i.e. the meta channel. Together, they could be encoded with all the information contained in the original light field, which is guaranteed by the autoencoder structure of the model in Fig.~\ref{fig:overview}.
Therefore, the meta channel plays the role of recording the original light field information complementary to the central view image, including: disparity between the central view and each other view, the occlusion content of other views from the central view, and other view-dependent information like non-Lambertian reflection. 
As illustrated in Figure~\ref{fig:representation}, the encoded meta channel resembles a depth map of the central view, with additional texture. This implies that apart from the depth/disparity information, other information is implicitly encoded in the form of structured patterns. To verify this, we conduct an ablation study in Section~\ref{subsec:ablation_study}, where the meta channel is replaced with a depth map of the central view and the reconstruction turns problematic.

\begin{figure}[!t]
	\centering
	\includegraphics[width=1.0\linewidth]{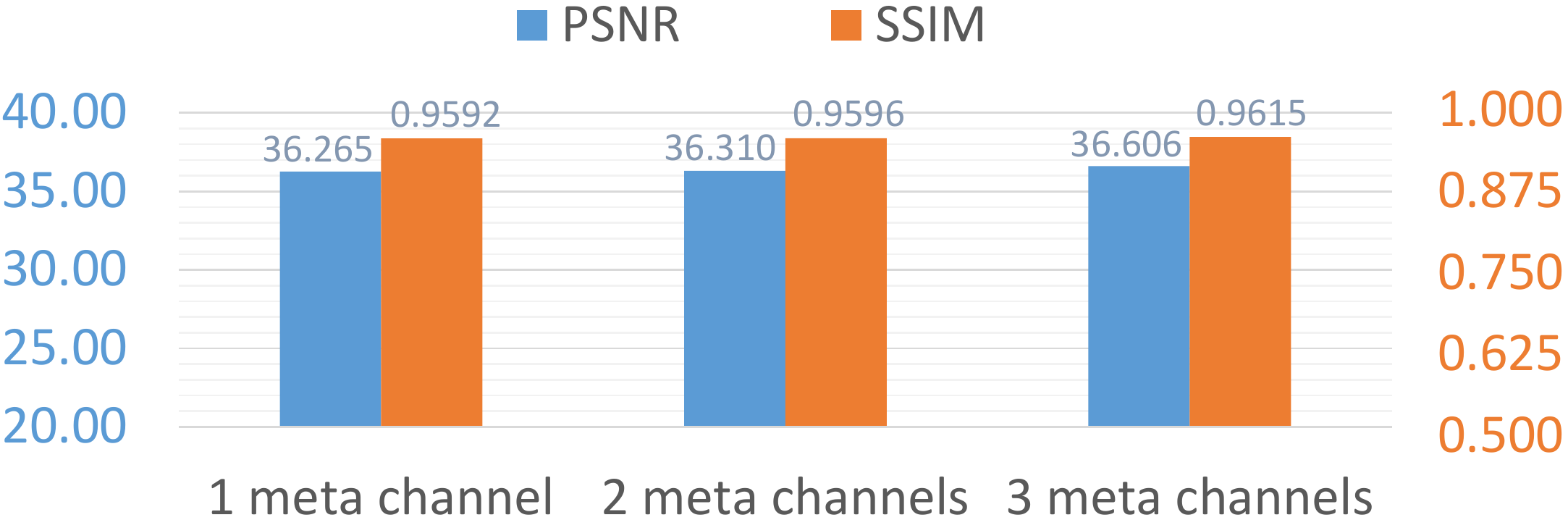}
	\caption{Statistical reconstruction accuracy against the meta channel numbers. This experiment is conducted over our testing dataset described in Section~\ref{sec:implementation}.}
 	\label{fig:meta_num}
\end{figure}

\begin{figure*}[!t]
  	\centering
  	\includegraphics[width=1.0\linewidth]{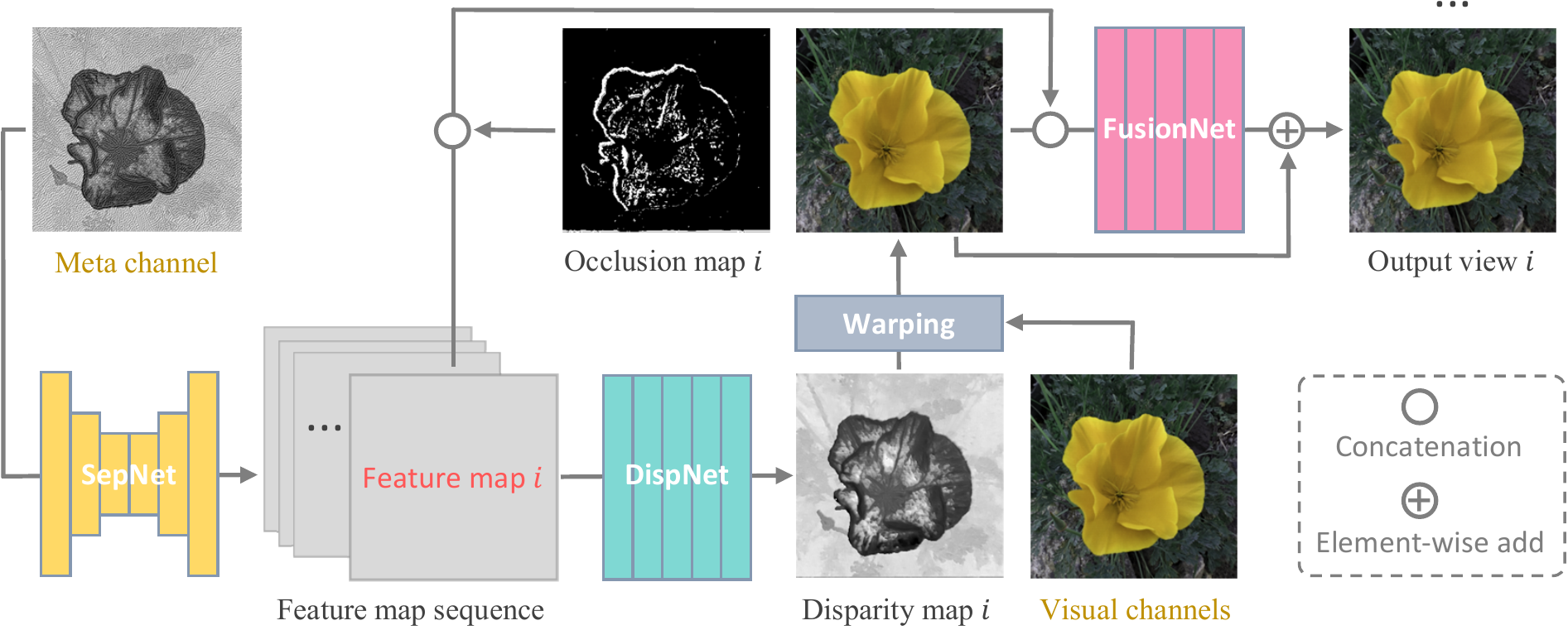}
  	\caption{A schematic illustration of the decoder. The input meta channel is decomposed into sequential features $\{\mathbf{F}_i\}_{i=1}^{M \times N}$ by the \textit{SepNet} module. \textit{DispNet} takes $\mathbf{F}_i$ to generate the corresponding disparity map $\mathbf{D}(\u_i)$. A warped view $\bar{\mathbf{I}}_i$ is generated from the disparity map and the visual channels.  Then, an occlusion mask is generated by checking inter-view disparity consistency, which it is fed along with $\bar{\mathbf{I}}_i$ and $\mathbf{F}_i$ to \textit{FusionNet}, to reconstruct the targeted view $i$.}
   	\label{fig:decoder}
\end{figure*}

Naturally, as input light field scale (e.g. spatial resolution or angular resolution) increases, the single meta channel may not have enough capacity to represent all those necessary information. In such situation, we could adaptively increase the number of meta channels.
Alternatively, the whole light field even could be encoded into the visual channels $\mathbf{I}^z$ directly via the encoder, i.e. $\mathbf{I}^z=\mathcal{E}(\lf)$, which means no meta channel is used. This is feasible to serves as a compact representation. However, the encoded visual channels are not tolerant to edits, because the implicitly encoded information may get damaged. So, separation between visual and meta channels is preferred in our targeted scenario.
In particular, for the case of $7 \times 7$ views in our dataset (detailed in Section~\ref{sec:implementation}), one meta channel achieves the best balance between efficiency and compactness, as the result shown in Fig.~\ref{fig:meta_num}.

\subsection{Editing-Aware Reconstruction}
\label{subsec:edit_reconstruction}
  
A U-shaped network is employed as the encoder, which learns a feature-domain representation of the input. However, it is more complex for the decoding process, which reconstructs a light field from an intermediate representation undergone potential edits. The naive solution of modeling the decoder with a single network, suffers from poor generalization to various edits to the visual channels (see Figure~\ref{fig:comp_regression}). Instead, we propose to break the decoder into several submodules: \textit{SepNet}, \textit{DispNet} and \textit{FusionNet}, each of which learns a specific task, as the schematic diagram shown in Figure~\ref{fig:decoder}. The detailed architectures of the encoder and decoder are provided in the supplementary material.

{\noindent\textbf{Feature Separation - \emph{SepNet}}.\quad}
As stated in Section~\ref{subsec:embeding_space}, the meta channel encodes all the information of the input light field, except for those in the visual channels. In order to reconstruct the light field views, we instruct the decoder to interpret that information as two categories: the inter-view disparity maps and other view-dependent information (e.g. occluded contents and non-Lambertian effects). Specifically, we utilize \textit{SepNet} module to decompose the meta channel $\mathbf{Z}$ into a sequence of feature maps $\{\mathbf{F}_i\}_{i=1}^{M \times N}$, where each $\mathbf{F}_i$ is self-contained and contains the exclusive information of view $i$. Then, each view $\mathbf{\tilde{I}}_i$ can be reconstructed from the feature map $\mathbf{F}_i$ and visual channels $\tilde{\view}_c$ independently.

{\noindent\textbf{Disparity Recovery - \emph{DispNet}}.\quad}
We extract from $\mathbf{F}_i$ the disparity information of view $i$, i.e. the disparity map $\mathbf{D}(\u_i)$, through \textit{DispNet}. Specifically, $\mathbf{D}(\u_i,\x)$ is a 2D vector, sliced from the 4D disparity field of $\lf$, which denotes the disparity of pixel $\lf(\u_i, \x)$ with respect to its horizontal and vertical neighboring views. Specifically, given view $\lf(\u_j)$ and the disparity map $\mathbf{D}(\u_i)$, we can obtain the warped view $i$ via:
\begin{equation}\label{eq:warp_formula}
\bar{\lf}(\u_i,\x) = 
\lf(\u_j, \x+(\u_j- \u_i)\cdot \mathbf{D}(\u_i, \x))
\end{equation}
So, once we have the disparity map $\mathbf{D}(\u_i)$, we can obtain the corresponding warped view by warping the visual channels $\view_c$ immediately. Obviously, the warped view $\bar{\lf}(\u_i)$ will not be exactly the same as our target view $\hat{\lf}(\u_i)$ because of occlusions and non-Lambertian effects. Following the idea of Ruder et al.~\cite{RuderGCPR16}, we can estimate the occlusion regions by performing a forward-backward consistency check of the disparity. In particular, when warping $\mathbf{I}_c$ to view $i$, the occlusion map $i$ can be approximated as:
\begin{equation}\label{eq:occlusion_formula}
\mathbf{O}(\u_i, \x) = ||\mathbf{D}(\u_i, \x) -\
\mathbf{D}(\u_c, \x+(\u_c - \u_i)\cdot \mathbf{D}(\u_i, \x))||_1
\end{equation}
where the greater the value of $\mathbf{O}(\u_i,\x)$, the more likely the sample $\lf(\u_i,\x)$ is occluded in $\bar{\view}_c$, thus indicating a stronger necessity to be corrected by \textit{FusionNet}.

\begin{figure}[!t]
	\centering
	\includegraphics[width=1.0\linewidth]{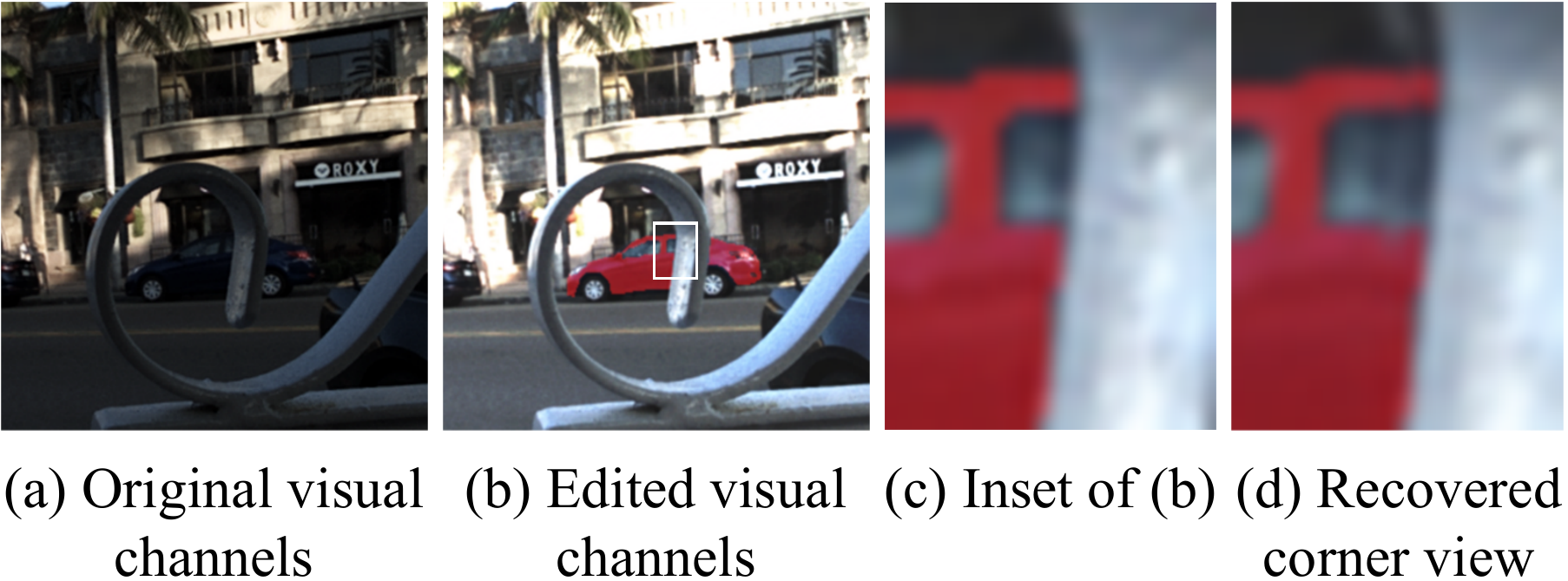}
	\caption{Occlusion recovery with editing effect propagated. Compared to the edited visual channels (c), the patch of the reconstructed corner view (d) reveals the occluded content with the editing effect reasonably propagated.}
 	\label{fig:context_synthesis}
\end{figure}

{\noindent\textbf{Context Based Synthesis - \emph{FusionNet}}.\quad}
To restore the occluded details and non-Lambertian effects, we further refine the warped view $\mathbf{\bar{I}}_i$ by \textit{FusionNet} through residual learning. 
Particularly, $\mathbf{F}_i$ provides the necessary view-dependent information of the original light field , while the occlusion map $\mathbf{O}_i$ indicates the problematic regions of the warped view $\mathbf{\bar{I}}_i$. 
Ideally, we would like to reconstruct the target view in a deterministic way. However, in the case of edited visual channels, it is substantially ambiguous how the edits should be propagated to those occluded regions of the target view.  
Actually, this issue depends on the context of the occlusion regions, and we instruct the model to learn these principles from large scale training pairs, i.e. $\{\tilde{\view}_c,\hat{\lf}\}$.
Whereas, it is impractical to manually prepare a diversely edited version of the light field dataset, as inter-view consistency is non-trivial to maintain. Instead, we propose to make use of some global editing operators $\mathcal{G}(\bullet)$, i.e. changing the hue, saturation, exposure, and contrast, whose parameters are sampled from an uniform distribution. The nice thing is that we can easily prepare the training pairs $\{\mathcal{G}(\mathbf{I}_c), \mathcal{G}(\mathbf{L})\}$ by applying these operators to the light field view individually, where the inter-view consistency can be preserved well. Notably, though only naive global color manipulation is involved during training, our model generalizes well to allowing many popular image edits to the visual channels, as described in Section~\ref{subsec:edit_evaluation}. The possible explanation is that these global edits and arbitrary local variant edits share the same essence of changing pixel values, and the model has to learn to propagate these changes to other views. Figure~\ref{fig:context_synthesis} evidences that our model can propagate the edited colors to those occluded areas through this strategy.

\subsection{Loss Function}
\label{subsec:loss_function}

The encoding and decoding subnetworks are jointly trained by minimizing a loss function that combines warping consistency loss $\mathcal{L}_W$, disparity regularity loss $\mathcal{L}_D$, and reconstruction loss $\loss_R$:
\begin{equation}\label{eq:loss_function}
\loss =  \omega_1 \loss_W + \omega_2 \loss_D + \omega_2 \loss_R
\end{equation}
where $\omega_1, \omega_2, \omega_3$ are the trade-off coefficients among different loss terms. We empirically set $\omega_1=0.5, \omega_2=0.01, \omega_3=1.0$ in our experiments.

First, the warping consistency loss measures the pixel-wise difference between the light field warped from the central view $\view_c$ and the input light field as:
\begin{equation}\label{eq: warping_loss}
\mathcal{L}_W =  \mathbb{E}_{\mathbf{L}_i \in \mathcal{S}}\{||\bar{\lf}_i - \lf_i||_1\}
\end{equation}
which actually constrains the disparity maps generated from \textit{DispNet}. The rationale lies in the fact that the baselines of light field cameras is very small, so the same object almost have the same appearance across different views. $||\bullet||_1$ means L1 norm and $\mathbb{E}_{\mathbf{L}_i \in \mathcal{S}}$ denotes the average operator over all the light fields in the training dataset $\mathcal{S}$. 

Only with $\loss_W$, the accuracy of disparity maps are still not guaranteed, because textureless regions are insensitive to incorrect disparity under $\loss_W$. It could be argued that errors in such regions are negligible if we only care about the correctness of the warped view. However, inaccurate disparity maps directly decrease the reliability of occlusion maps, thus complicates the training of \textit{FusionNet} which takes as input the occlusion map as correction hints. So, we explicitly strengthen the disparity consistency between neighbor views with the regularity loss:
\begin{equation}\label{eq: disparity_loss}
\mathcal{L}_D =  \mathbb{E}_{\mathbf{L}_i \in \mathcal{S}}\{||\mathbf{D}(\u_i,\x)- \mathbf{D}(\u_i-\mathbf{1},\x + \mathbf{D}(\u_i,\x))||_1\}
\end{equation}
which encourages both the predicted disparity to be consistent across views and the occluders to be sparse. For efficiency, the disparity consistency of each view is only checked with three neighbor views, i.e. $\textbf{1} = \{(0,1), (1,0), (1,1)\}$. Under this constraints, the light field views are related as a connected graph, where each view is affected by the rest indirectly. 

Finally, we regulate the reconstructed light field $\tilde{\lf}$ to be the same as the target one $\hat{\lf}$ via the reconstruction loss:
\begin{equation}\label{eq: resonstruct_loss}
\mathcal{L}_R =  \mathbb{E}_{\lf_i \in \mathcal{S}}\{\alpha||\tilde{\lf}_i - \hat{\lf}_i||_1+\beta ||\text{SSIM}(\tilde{\lf}_i, \hat{\lf}_i)||_1\}
\end{equation}
where $\text{SSIM}(\bullet, \bullet)$ measures the average SSIM~\cite{WangTIP04} over all the views of light fields.
$\alpha=1.0$ and $\beta=0.02$ are used to balance the magnitude between the two loss terms. We found SSIM helps achieving more accurate details, especially in the disoccluded regions.

\section{Implementation Details}
\label{sec:implementation}

{\noindent\textbf{Dataset}.\quad}
Our dataset is built from two publicly available datasets: the Stanford Lytro Lightfield Archive~\cite{StanfordArchive16} and the dataset from Kalantaru et al.~\cite{KalantariTOG16}, both of which were captured with Lytro Illum cameras. In particular, 304 light fields from~\cite{StanfordArchive16} and 102 light fields from~\cite{KalantariTOG16} compose our dataset, with repeated or low-quality samples manually identified and removed. Each light field has $376 \times 541$ spatial samples and $14 \times 14$ angular samples, but only the central $7 \times 7$ grid of angular samples are used in our experiments because many peripheral angular samples are outside the camera aperture.
Among the 406 light fields, 326 ones were randomly selected as the training dataset and the remaining 80 were used as the testing dataset. In the training dataset, each light field exemplar, containing 49 RGB views, is randomly cropped into 20 patches of $128 \times 128$ pixels. Also, we augment each patch by flipping it horizontally or rotating it $180^{\circ}$, where the image order are adjusted accordingly to preserve the epipolar geometry of the light field. Finally, with textureless patches filtered out by an automatic procedure computing average gradient, we get $14,447$ light field samples to train our model.

{\noindent\textbf{Training}.\quad}
As stated in Section~\ref{subsec:loss_function}, both the encoding subnetwork and decoding subnetwork are trained jointly. However, we found that directly training the whole networks from scratch suffers from low-speed convergence. This is explained by the fact that at the beginning, the disparity maps from \textit{DispNet} are mostly random noise, which thus causes problematic occlusion maps, confusing the training of \textit{FusionNet}. 
To avoid this situation, we propose to train the network in two stages. In the first stage, all the networks excluding \textit{FusionNet}, are trained jointly for the first $10,000$ iterations, so as to warm up the \textit{DispNet}. In the second stage, all the networks including \textit{FusionNet} are trained for another $50,000$ iterations. Specifically, we optimize our model using the Adam optimizer~\cite{Diekerik2014}, with the learning rate fixed as $0.0002$ in the first stage and linearly decreased to its $1.0\%$ in the second stage.

We implemented our method using Pytorch\footnote{Pytorch: \url{https://pytorch.org/}}, a Python-based open-source deep learning framework. All the experiments were ran on a server equipped with two NVIDIA Geforce GTX 1080 Ti.
The training consumes roughly 48 hours. A light field with $7 \times 7$ angular samples for $376 \times 541$ spatial samples takes about $0.043$s (seconds) to encode the meta channel, and $2.056$s to reconstruct a full light field ($0.042$s per view) from it.
\textit{The source code and trained models will be made public upon publication}.

%% file: Results.tex
\section{Experimental Results}
\label{sec:experiment}

The propose method is evaluated in three aspects: reconstruction accuracy, editing propagation consistency, and ablation study on key designs. Additional results and video visualizations are available in the supplementary material.

\begin{table}[!t]
	\centering
	\renewcommand{\tabcolsep}{7pt}
	\caption{Reconstruction evaluation on our testing dataset (central $5 \times 5$ views used). Higher PSNR/SSIM is better.}
	\begin{tabular}{ccccc}
		\hline
		\multirow{2}{*}{Method}      & \multicolumn{2}{c}{PSNR}       & \multicolumn{2}{c}{SSIM}    \\
		\cline{2-5}                              & Mean           & Stddev                 & Mean   			& Stddev	  	  \\ \hline
	    Inagaki et al~\cite{InagakiECCV18}   & 32.676		& 0.8508    & 0.9211		      & 0.0095		    \\
		Ours          	                             & \textbf{37.232}	     & 2.0698    & \textbf{0.9638}             & 0.0083          \\ \hline
	\end{tabular}
	\label{tab:restoration_comp1} 
\end{table}

\begin{figure}[!t]
	\centering
	\includegraphics[width=1.0\linewidth]{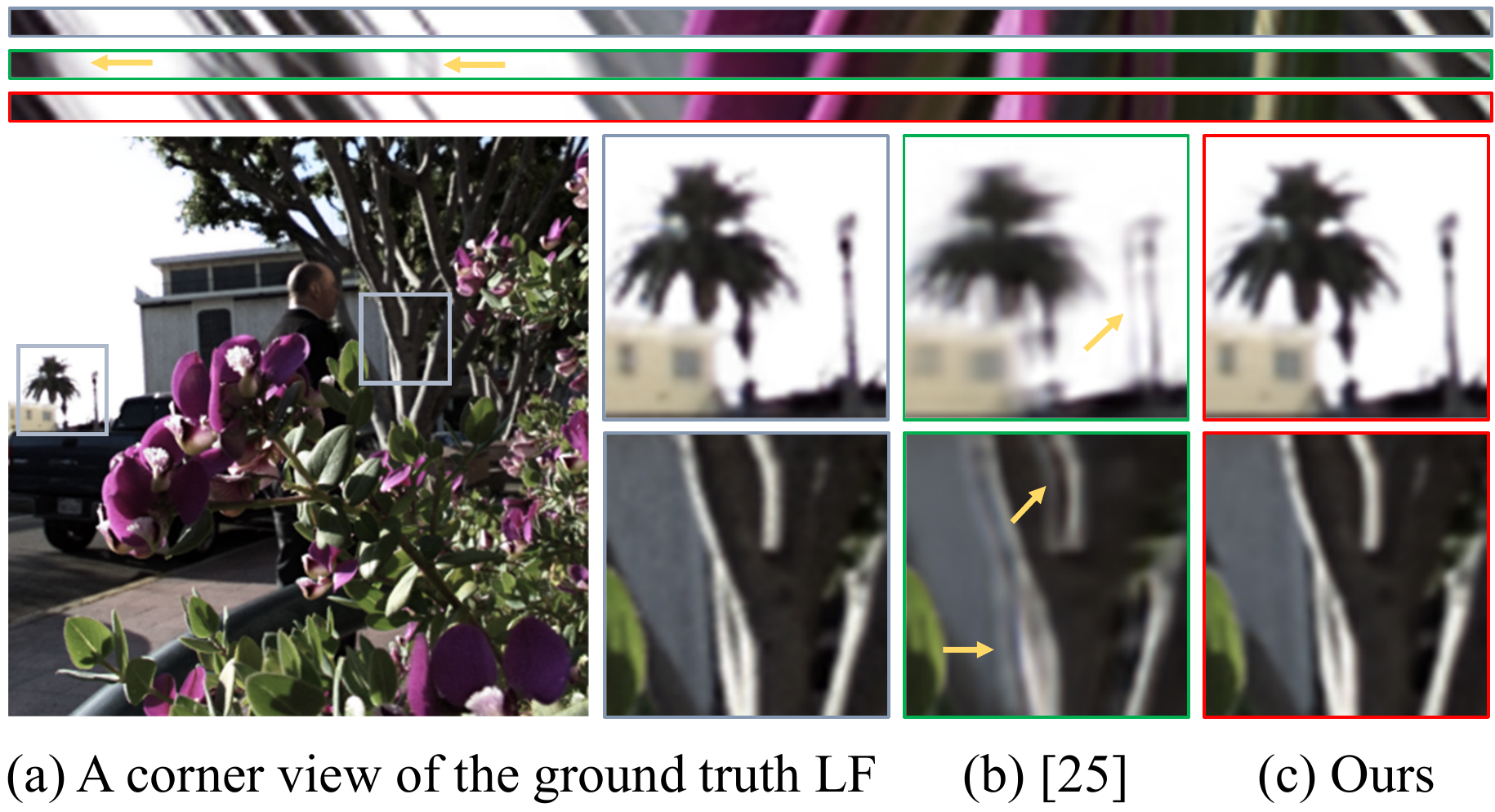}
	\caption{Visual comparison with~\cite{InagakiECCV18}. The insets show the reconstruction details while the EPIs evaluate both the inter-view consistency and parallax correctness. The boundary color of inset represents different methods (grey denotes the ground truth), and yellow arrows indicate problematic regions.}
 	\label{fig:comp_inaga}
\end{figure}

\subsection{Reconstruction Accuracy}
\label{subsec:restoration_evaluation}

We evaluate the reconstruction accuracy of our representation when no edits are performed. The peak signal-to-noise ratio (PSNR) and structural similarity (SSIM) are used as measurement metrics. As a compact light field representation that is instantly viewable, our method is firstly compared with two existing baselines: Inagaki et al.~\cite{InagakiECCV18} represent a light field as one or few coded images by learning micro-lens weighting masks; Srinivasan et al.~\cite{SrinivasanICCV17} synthesize a light field from a regular 2D image. In any case, we would like to clarify that those methods were devised for different purposes, so more than to compare with them,  we use them to highlight the value of our representation.

%
\begin{table}[!t]
	\centering
	\renewcommand{\tabcolsep}{6pt}
	\caption{Reconstruction evaluation on Flowers dataset~\cite{SrinivasanICCV17} ($7 \times 7$ views). Higher PSNR/SSIM is better.}
	\begin{tabular}{ccccc}
		\hline
		\multirow{2}{*}{Method}      & \multicolumn{2}{c}{PSNR}       & \multicolumn{2}{c}{SSIM}    \\
		\cline{2-5}                              & Mean           & Stddev                 & Mean   			& Stddev	  	  \\ \hline
	   Srinivasan et al~\cite{SrinivasanICCV17}   & 33.365		& 3.2325    & 0.9115		      & 0.0380		    \\
		Ours          	           & \textbf{40.944}	       & 1.8097       & \textbf{0.9719}             & 0.0062          \\ \hline
	\end{tabular}
	\label{tab:restoration_comp2}
\end{table}

\begin{figure}[!t]
	\centering
	\includegraphics[width=1.0\linewidth]{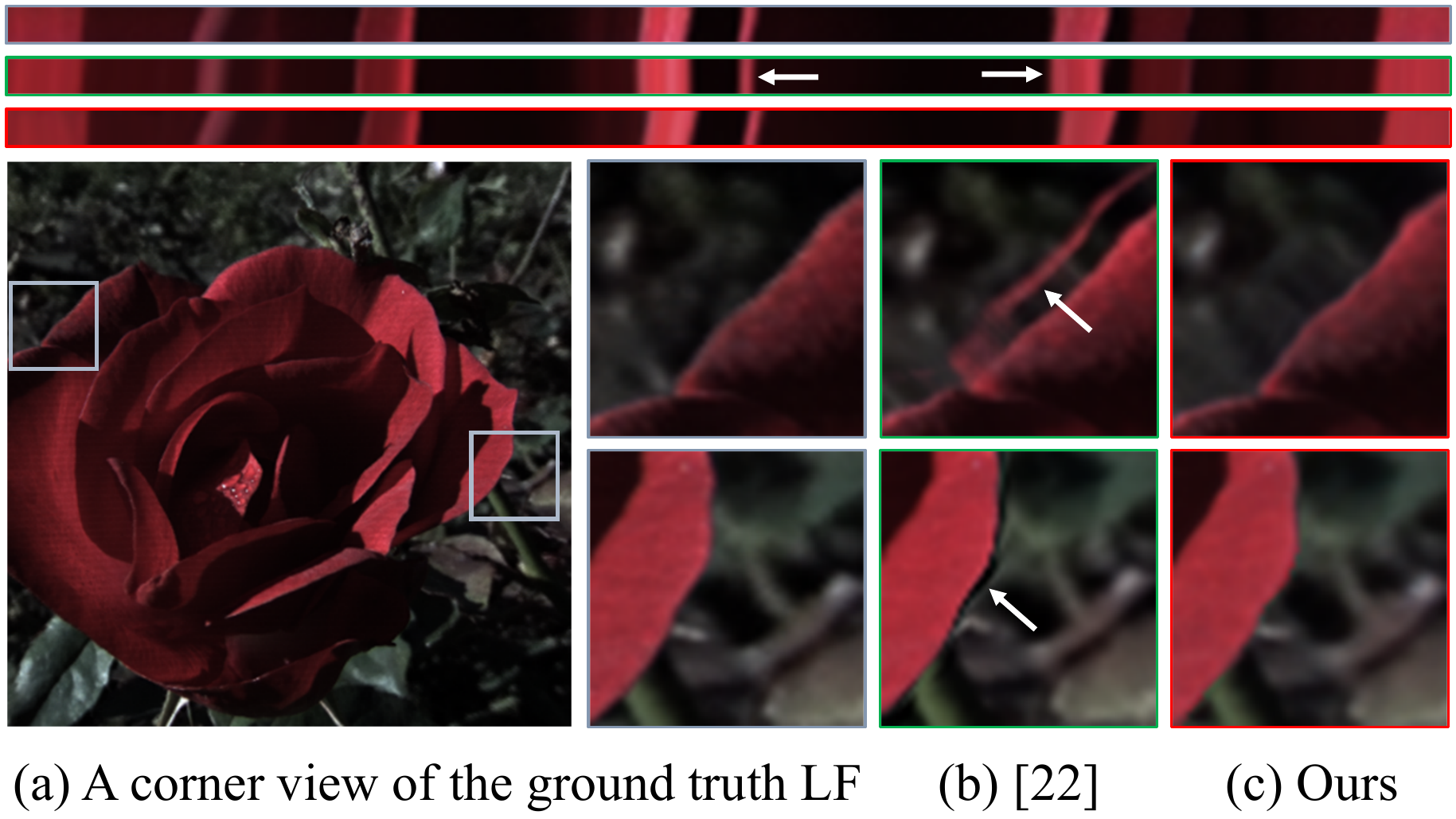}
	\caption{Visual comparison with~\cite{SrinivasanICCV17}. The insets show the reconstruction details while the EPIs evaluate both the inter-view consistency and parallax correctness. The boundary color of inset represents the different methods (grey denotes the ground truth), and white arrows indicate problematic regions.}
 	\label{fig:comp_srini}
\end{figure}

\begin{figure*}[!t]
	\centering
	\includegraphics[width=1\linewidth]{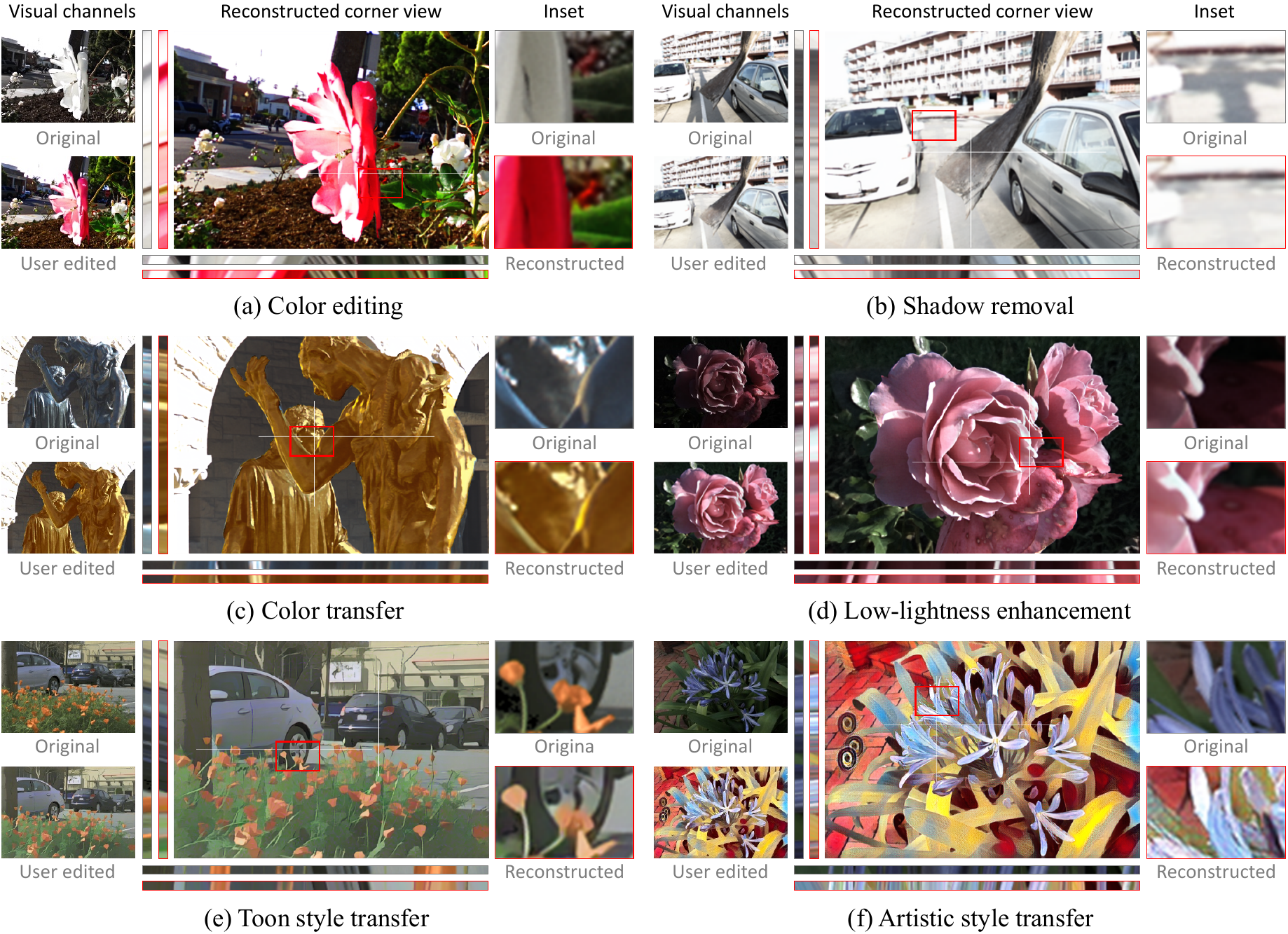}
	\caption{Showcase of light field reconstruction from the visual channel with typical edits applied (via interactive edit with \textit{Photoshop} or popular image processing algorithms). The attached horizontal and vertical EPIs (for the corresponding scanlines highlighted in white) visualize the inter-view consistency and parallax correctness w.r.t that of the original light field.}
 	\label{fig:showcase_edit}
\end{figure*}

To compare with~\cite{InagakiECCV18}, we use the pre-trained model provided by the authors, which reconstructs a light field of $5 \times 5$ views from two coded images. Since our model is trained for light fields of $7 \times 7$ views, we simply take the central $5 \times 5$ views of the reconstructed light fields for comparison. The results are presented in Table~\ref{tab:restoration_comp1}, which shows our method outperforms~\cite{InagakiECCV18} at a large margin. The main reason is that~\cite{InagakiECCV18} regulates the encoding scheme as polynomial combination of the input light field views, which significantly limits the solution space. On the contrary, our encoding scheme is adaptively learned under the guidance of reconstruction accuracy. To visualize the reconstructed light field, we present a corner view along with epipolar slices in Figure~\ref{fig:comp_inaga}. It is worth noting that as encoded with information, this 2-image representation proposed in~\cite{InagakiECCV18} cannot be directly edited.

To compare with~\cite{SrinivasanICCV17}, we download the source code and dataset from the author's website and train it from scratch. Since synthesizing a light field from a single image is extremely ill-posed, the model trained on specific type of scene usually generalize poorly to new types of scenes. So for fairness, we perform comparisons over the dataset used in their paper, which consists of 100 light fields of flowers. The quantitative results are tabulated in Table~\ref{tab:restoration_comp2}, and a typical example for visual comparison is illustrated in Figure~\ref{fig:comp_srini}. As expected,~\cite{SrinivasanICCV17} can not recover accurate details that are occluded in the input single image, so they are filled with dark pixels or surrounding textures instead. However, it is fair to say that this method was not meant to be used as a compact representation for captured light fields, but a solution to synthesize plausible light fields from regular 2D images.

\begin{figure*}[!t]
	\centering
	\includegraphics[width=1\linewidth]{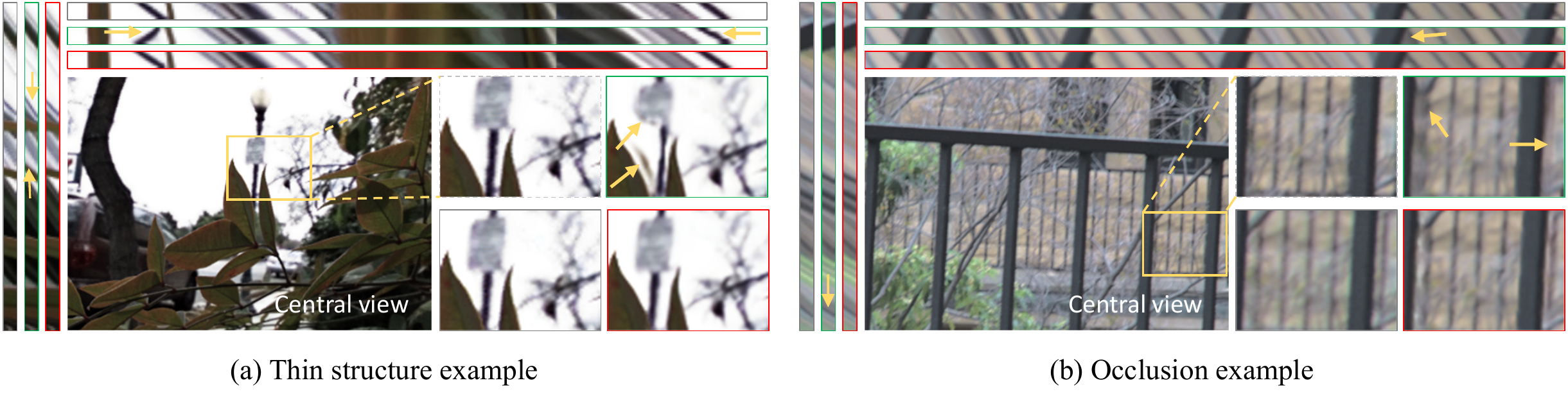}
	\caption{Reconstruction comparison between RGB-D and our representation. The insets with solid boundaries show the reconstructed corner views and EPIs of different methods: \textbf{grey} denotes the ground truth, \textbf{green} denotes RGB-D baseline, and \textbf{red} denotes ours. The yellow arrows indicate problematic regions.}
 	\label{fig:comp_rgbd}
\end{figure*}

\subsection{Editing Propagation}
\label{subsec:edit_evaluation}

Our proposed light field representation allows edits on the visual channels, as long as they do not alter the underlying scene geometry. Although this leaves out basic edits like scaling, copying or inpainting, we believe the proposed pipeline already supports plenty of advanced and useful edits.
Example of edits supported include: global/local color manipulation (e.g. exposure, contrast, saturation, hue, etc), object texture modification, and most image processing algorithms (e.g. color transfer, denoising, dehazing, shadow removal, style transfer, etc).

Firstly, we evaluate the propagated edits by visually inspecting the inter-view consistency and parallax correctness.
In Fig.~\ref{fig:showcase_edit}, the light fields are reconstructed from the visual channels that are edited through global and local color adjustment with \textit{Photoshop} (a), shadow removal algorithm~\cite{CunAAAI20}, photorealistic color transfer~\cite{DYooICCV19}, low-lightness enhancemen~\cite{WangCVPR19}, photo cartoonization~\cite{WangCVPR20}, and style transfer technique~\cite{JustinECCV16} respectively.
For each example, we illustrate one of the challenging corner views that have the largest disparities with respect to the central view, and the horizontal and vertical epipolar plane images (EPIs) of spatial segments, with comparison to the original light fields.
By comparing the EPIs, we observe that the reconstructed light fields keep a very similar geometric structure, i.e. the parallax captured by the slopes of the color strips, to the original light fields. Besides, the insets of the corner view illustrate the reconstruction quality in local regions, where no artifacts or visual inconsistencies are observed.
Anyhow, the toon style transfer simplifies structural details while the artistic style transfer tends to introduce extra textures to the results, and thus they may cause some reasonable inconsistency between the EPIs of the original light fields and the edit propagated ones.
Note that these light fields were edited through the visual channels, using existing 2D image editing software. This is not only much more memory-friendly in run time, but allows complex edits not possible with previous light field editing tools. Readers are recommended to check out video materials for these visual comparison. Also, more results are available in the supplementary material.

In the absence of ground-truth edited versions of our test scenes, we evaluate the edit propagation quality quantitatively in two aspects. Firstly, as one of the supported edits, global color operations, i.e. changing the hue, saturation, exposure, and contrast, are adopted to quantitatively test our method over the whole testing dataset. In particular, we randomly apply one of these operations (sampling the parameters from a uniform distribution) to the views of each light field individually, which can approximately serve as the ground-truth edited version. The average PSNR and SSIM are $36.099$ and $0.9539$ respectively, just slightly lower than the reconstruction accuracy without edits involved (i.e. $36.265$ and $0.9592$).
However, global color modification is just one of the edits supported by our method. For quantitative evaluation on more diverse edits, e.g. those demonstrated in Fig.~\ref{fig:showcase_edit}, we propose to utilize an indirect measurement: given a light field $\lf$ and its edited visual channels $\tilde{\view}_c$, we run the encoding and decoding recurrently for twice:
$\lf^{rec}=\mathcal{D}(\mathcal{E}(\tilde{\lf}), \view_c)$ with $\tilde{\lf}=\mathcal{D}(\mathcal{E}(\lf), \tilde{\view}_c)$. Then, we can measure the difference between the recurrently reconstructed light field $\lf^{rec}$ and the original light field $\lf$ conveniently. The results are tabulated in Table~\ref{tab:edit_evaluation}, with a control group that has $\tilde{\view}_c=\view_c$ provided for reference. It shows that our edit reconstruction almost has the same accuracy as the reconstruction without any edit involved, which suggests the good edit propagation quality in some sense.

\begin{table}[!t]
	\centering
	\renewcommand{\tabcolsep}{5pt}
	\caption{Edit propagation evaluation on typical examples shown in Fig.~\ref{fig:showcase_edit}. The control group with no edits applied is provided for reference. Higher PSNR is better.}
	\begin{tabular}{ccccc}
		\hline
		\multirow{2}{*}{Example}    & \multicolumn{2}{c}{With Edit}     & \multicolumn{2}{c}{No Edit} \\
		\cline{2-5}                               & PSNR         & SSIM                          & PSNR   		           & SSIM	  	   \\ \hline
		Color edit                                 & 33.068       & 0.9733                      & 33.144	  	          & 0.9736      \\
	    Shadow removal                     & 34.159       & 0.9748                      & 34.226		          & 0.9752		\\
		Color transfer          	            & 35.509	    & 0.9726                     & 35.538                 & 0.9726      \\
	    Lightness enhance                   & 35.342       & 0.9600                      & 35.504		          & 0.9619       \\
	    Toon style transfer                  & 41.092       & 0.9873                      & 41.185		           & 0.9861       \\
	    Artistic style transfer               & 39.865       & 0.9755                      & 40.109		            & 0.9761       \\ \hline
	    \textbf{Average}                      & \textbf{37.663} &\textbf{0.9739} & \textbf{37.790}  	& \textbf{0.9742}  \\ \hline
	\end{tabular}
	\label{tab:edit_evaluation}
\end{table}

\subsection{Ablation Study}
\label{subsec:ablation_study}

\begin{table}[!t]
	\centering
	\renewcommand{\tabcolsep}{6pt}
	\caption{Quantitative comparison with baselines. The reconstruction accuracy is compared over the testing dataset, while the editing propagation is compared over the edit cases shown in Fig.~\ref{fig:showcase_edit}. Higher PSNR/SSIM is better.}
	\begin{tabular}{ccccc}
		\hline
		\multirow{2}{*}{Method}    & \multicolumn{2}{c}{Reconstruction}    & \multicolumn{2}{c}{Edit Propagation}    \\
		\cline{2-5}                             & PSNR               & SSIM                              & PSNR   			       & SSIM	  	   \\ \hline
	    RGB-D baseline                      & 34.139             & 0.9454                         & 34.541		            & 0.9657	   \\
	    Regression baseline  	         & \textbf{36.505}  & \textbf{0.9620}      & 36.143                  & 0.9723       \\ 	   	
	    Ours                                        & 36.265		     & 0.9592            & \textbf{37.663}	 & \textbf{0.9739}      \\\hline
	\end{tabular}
	\label{tab:ablation_comp}
\end{table}

{\noindent\textbf{Meta Channel}.\quad}
We stated the meta channel encodes all the necessary information that enables the original light filed to be reconstructed deterministically. However, one could argue that the meta channel might be no more than a depth map and all the other missing information relies on the prediction by CNN. To study this issue, we constructed a baseline that replace our learned meta channel with a depth map of the central view, namely the RGB-D representation, for comparative analysis. Particularly, the depth map is estimated from the whole light field through the state-of-the-art method~\cite{WangICCV15}. To ensure all the other conditions unchanged, we utilize a CNN with the same architecture as our decoder, to reconstruct the light field from such RGB-D representation under the same loss function in Eq.~\ref{eq:loss_function}.
The quantitative results are tabulated in Table~\ref{tab:ablation_comp}, showing that the reconstruction accuracy of RGB-D is inferior to ours. However, the margin is not very large, because the narrow baseline of light field cameras avoids most inter-view occlusions. Anyhow, the qualitative results illustrated in Fig.~\ref{fig:comp_rgbd} reveal the major limitation of the RGB-D baseline with respect to our proposed representation.

\begin{figure}[!t]
	\centering
	\includegraphics[width=1\linewidth]{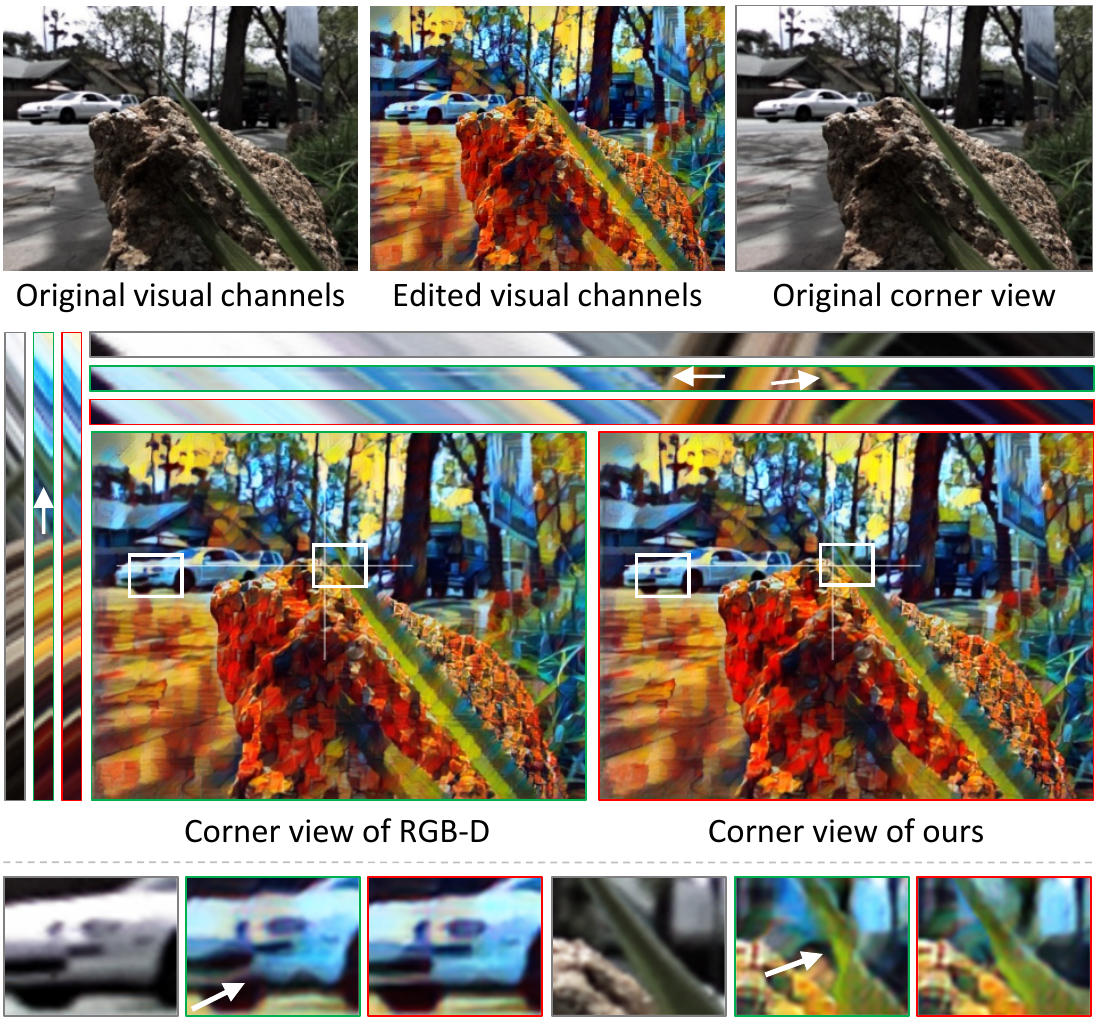}
	\caption{Edit propagation comparison between RGB-D and our representation. The insets with color boundaries show the reconstructed corner views and EPIs of different methods: \textbf{grey} denotes the ground truth without edits, \textbf{green} denotes RGB-D baseline, and \textbf{red} denotes ours. The white arrows indicate problematic regions.}
 	\label{fig:comp_rgbd_edit}
\end{figure}

Firstly, the reconstruction accuracy of RGB-D highly relies on the quality of the depth map, however depth estimation from light field is challenging itself, especially in those busy background regions with fine structures. As a result, distortion and ghost effect appeared in the inset of Fig.~\ref{fig:comp_rgbd}(a). In theory, the depth map is unable to record the accurate depth of object boundaries due to the finite resolution, i.e. pixels of object boundary tends to be a mixture of foreground and background, and are associated with multiple depths. Therefore, it makes no sense to take an explicit depth map for information representation. In contrast, our meta channel is adaptively learned through an auto-encoder framework as a unrestricted feature map. It is flexible enough to encode any necessary information in order to reconstruct the original light field, including those occlusions, which in contrast is impossible for a depth map. In Fig.~\ref{fig:comp_rgbd}(b), the occluded baluster is recovered from our representation while still missing in the result of RGB-D baseline.
Naturally, these limitations of RGB-D baseline retain in the editing propagation results, as demonstrated in Fig.~\ref{fig:comp_rgbd_edit}.

\begin{figure}[!t]
	\centering
	\includegraphics[width=1.0\linewidth]{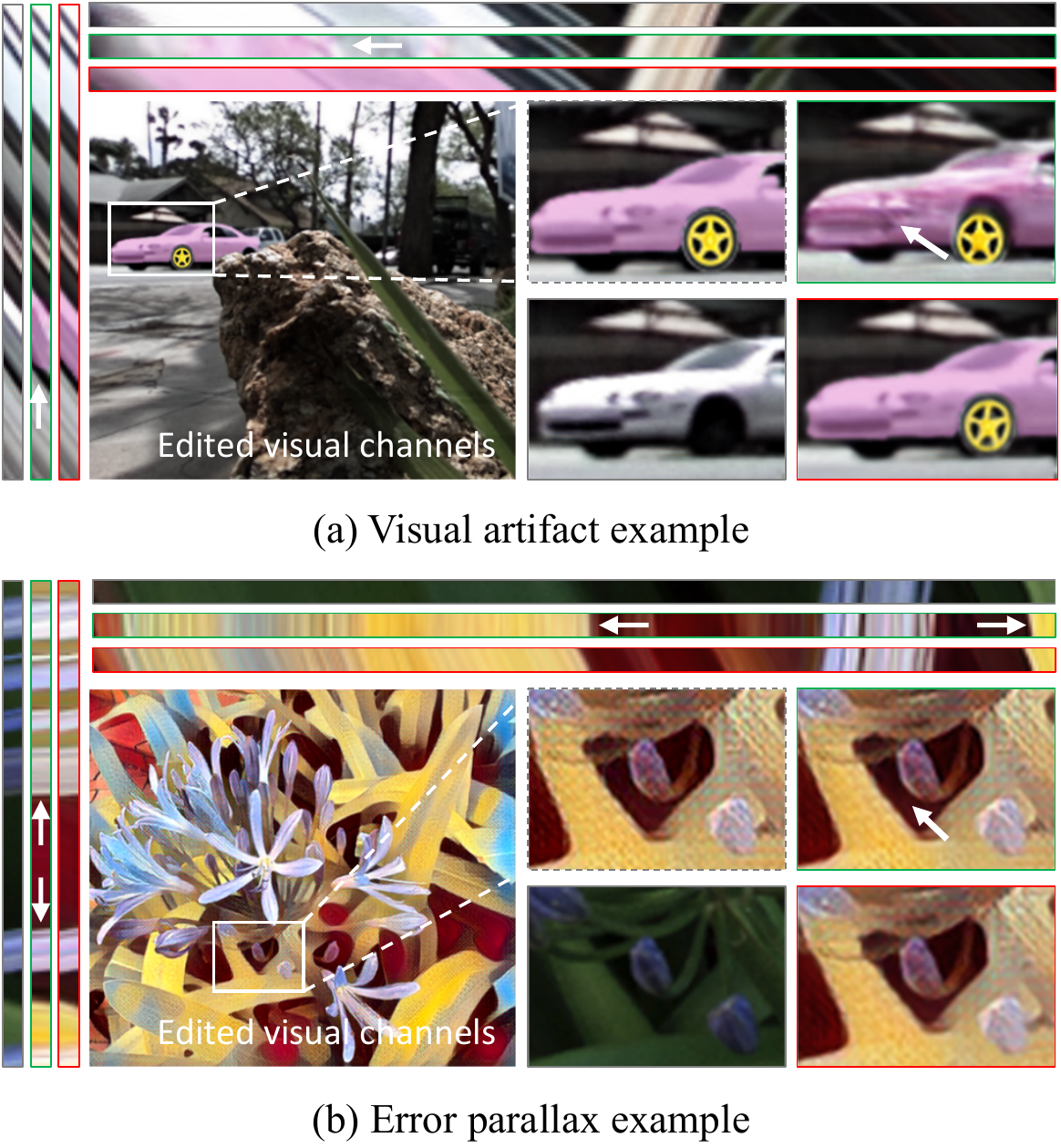}
	\caption{Edit propagation comparison between different decoding networks. The insets with solid boundaries show the reconstructed corner views and EPIs of different methods: \textbf{grey} denotes the ground truth without edits, \textbf{green} denotes regression baseline, and \textbf{red} denotes ours. The white arrows indicate problematic regions.}
 	\label{fig:comp_regression}
\end{figure}

{\noindent\textbf{Decoding Network}.\quad}
Our proposed decoding network includes three specific subnetworks. To show its necessity, we compare it with an intuitive baseline, i.e. \textit{Regression} which uses a single network (the same as the encoder) to reconstruct the target light field from the generated representation directly. The quantitative results are tabulated in Table~\ref{tab:ablation_comp}. Regression baseline achieves the highest reconstruction accuracy, because of the more flexible decoding mechanism that runs in the unrestricted feature domain completely. However, we find that it is less compatible to edits to the visual channels, especially for those with significantly changed colors or textures.
Fig.~\ref{fig:comp_regression} illustrates two typical examples. Texture replacement causes noticeable visual artifacts and parallax failure. Severe texture modification, e.g. by artistic style transfer, makes the reconstructed views almost present zero parallax, as EPIs shown in Fig.~\ref{fig:comp_regression}(b). Also, Table~\ref{tab:ablation_comp} tabulates the quantitative results on the edit cases exampled in Fig.~\ref{fig:showcase_edit}. 
Comprehensively, our proposed decoding network enables our representation to reconstruct artifacts-free views with edits propagated consistently.

\subsection{Refocusing Application}
\label{subsec:application}

\begin{figure}[!t]
	\centering
	\includegraphics[width=1.0\linewidth]{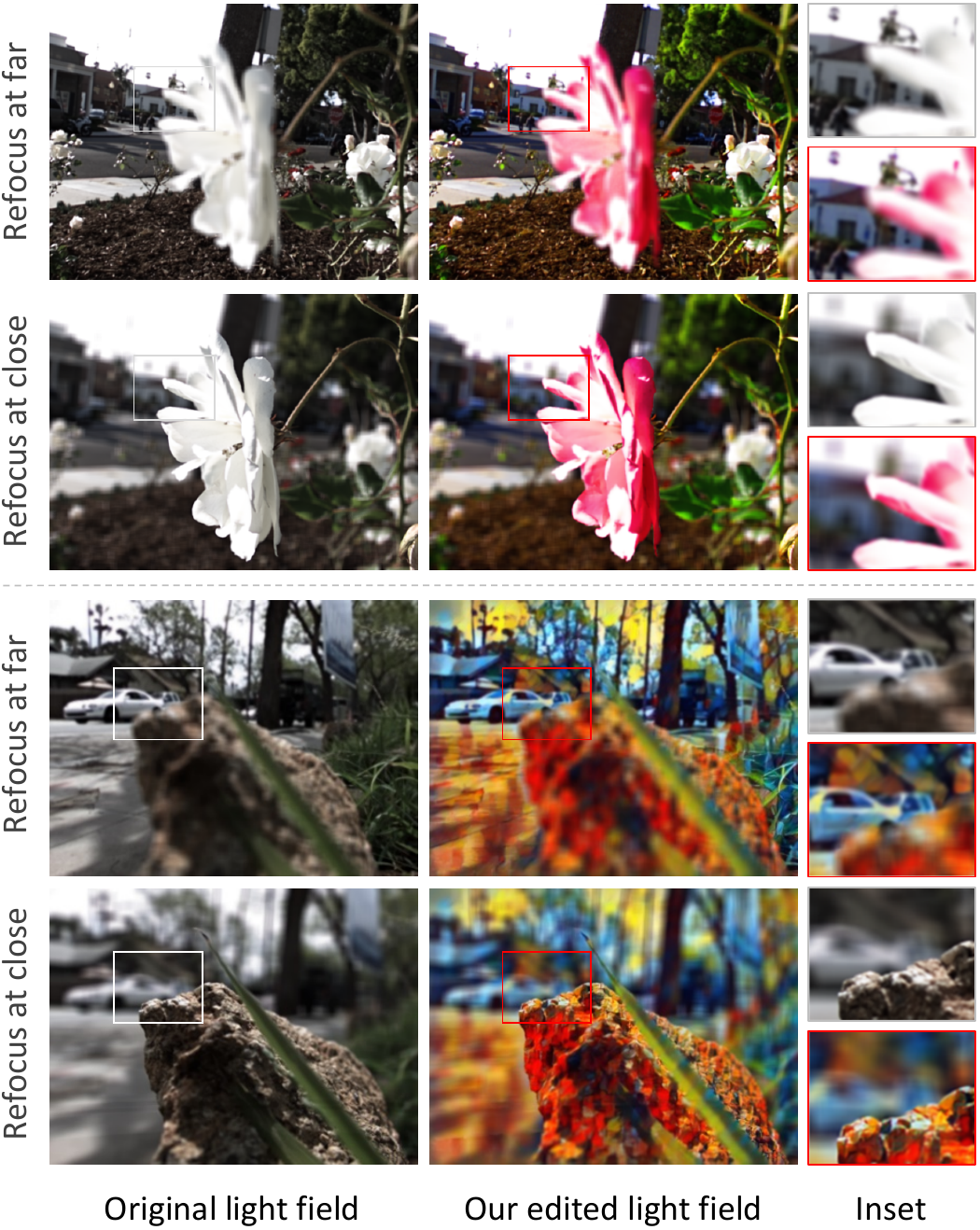}
	\caption{Synthetic refocusing comparison between the original and the edited light fields. For each example, the results of focusing at close and far distance are illustrated. The insets offer clearer inspection.}
 	\label{fig:refocus_app}
\end{figure}

One typical application of light fields is computational refocusing. Since our method enables lots of fancy 2D image processing algorithms applicable to light fields, it is interesting to utilize those edited light fields for refocusing application. Fig.~\ref{fig:refocus_app} demonstrates two examples, each of which presents both close and far focusing results. By comparing to the original light fields, our reconstructed light fields achieved comparable refocusing quality. 
In practice, our proposed technique may even benefit to artistic creation. As it is not easy to manually simulate realistic refocusing effects, our technique enable to demonstrate physically correct refocusing effects on cartoon photos, artistic photos.

\subsection{Discussion and Limitation}
\label{subsec:limitation}

Our method is the first attempt to allow light field editing based on statistical learning. The proposed separation of visual channels and meta channel offers desirable flexibility. First, the visual channels represents the visual content of the light field as a normal RGB image, which can be easily edited by the well established 2D image algorithms or softwares. Second, the meta channel records those complimentary information as an unrestricted feature map, and its channel number is adjustable according to the light field scale. Compared to a depth map, our meta channel can be encoded with various information beyond depth, which enables higher representation accuracy and better generalization to light fields of different scales.

Although we already demonstrated its potential through complex and useful edits, other edits affecting the underlying geometry and materials of the scene require further study. For example, currently compositing or removing a foreground object on the visual channels could cause distortion to the reconstructed views, because of the mismatched visual contents with the scene geometry.
Future work is to allow the meta channel updating adaptively to match the edited visual channels during the decoding phase. We expect more researches to enrich the functionality of the proposed representation framework.